\documentclass[letterpaper, 10 pt, conference]{ieeeconf}  % Comment this line out if you need a4paper

\IEEEoverridecommandlockouts                              % This command is only needed if 
\usepackage{cleveref}

\title{\LARGE \bf
Learning to Swim: Reinforcement Learning for 6-DOF Control of Thruster-driven Autonomous Underwater Vehicles
}

\author{Levi Cai$^{1*}$, Kevin Chang$^{2*}$, and Yogesh Girdhar$^{3}$% <-this % stops a space
\thanks{*These authors contributed equally to this work}% <-this % stops a space
\thanks{$^{1}$Levi Cai is with the Massachusetts Institute of Technology Woods Hole Oceanographic Institution Joint Program
        {\tt\small cail@mit.edu}}%
\thanks{$^{2}$Kevin Chang is with the Oregon State University {\tt\small changk2@oregonstate.edu}}%
\thanks{$^{3}$Yogesh Girdhar is with the Woods Hole Oceanographic Institution
        {\tt\small ygirdhar@whoi.edu}}%
}

\begin{document}

\maketitle
\thispagestyle{empty}
\pagestyle{empty}

%%%%%%%%%%%%%%%%%%%%%%%%%%%%%%%%%%%%%%%%%%%%%%%%%%%%%%%%%%%%%%%%%%%%%%%%%%%%%%%%
\begin{abstract}

Controlling AUVs can be challenging because of the effect of complex non-linear hydrodynamic forces acting on the robot, which are significant in water and cannot be ignored. The problem is exacerbated for small AUVs for which the dynamics can change significantly with payload changes and deployments under different hydrodynamic conditions.  The common approach to AUV control is a combination of passive stabilization with added buoyancy on top and weights on the bottom, and a PID controller tuned for simple and smooth motion primitives. However, the approach comes at the cost of sluggish controls and often the need to re-tune controllers with configuration changes. In this paper, we propose a fast (trainable in minutes), reinforcement learning-based approach for full 6 degree of freedom (DOF) control of a thruster-driven AUVs, taking 6-DOF command-conditioned inputs direct to thruster outputs. We present a new, highly parallelized simulator for underwater vehicle dynamics. We demonstrate this approach through zero-shot sim-to-real (with no tuning) transfer onto a real AUV that produces comparable results to hand-tuned PID controllers. Furthermore, we show that domain randomization on the simulator produces policies that are robust to small variations in vehicle's physical parameters.

%Reinforcement learning is a promising field of research that seeks to learn a model for robotic control over experiences collected by interactions between the robot and its environment. Due to the high cost of collecting data in the real-world, highly parallelized simulation-based model training, which can then be transferred to real vehicles without additional training, have shown remarkable promise. However, existing simulation frameworks for underwater vehicles are computationally expensive and fail to realistically model hydrodynamic conditions, resulting in poor real-world transfer. We propose an underwater simulation that enables simulating thousands of vehicles at a time and uses domain randomization to train policies that are robust to variations in physical parameters. We show how our simulation can be used to evaluate the performance of different policies across various environments. We also demonstrate that using our simulation, we are able to perform zero-shot transfer into the real-world with performance comparable to pure PID control.

\end{abstract}

%%%%%%%%%%%%%%%%%%%%%%%%%%%%%%%%%%%%%%%%%%%%%%%%%%%%%%%%%%%%%%%%%%%%%%%%%%%%%%%%
\section{INTRODUCTION}

Autonomous underwater vehicles (AUVs) and Remotely Operated Vehicles (ROVs) are being used in increasingly challenging environments and tasks, such as operating in harsh conditions under ice \cite{jakuba2024four} to actively tracking highly dynamic animals such as sharks \cite{hawkes2020autonomous}, which require highly capable control systems.

Currently, there are various control methodologies for AUVs, including PID, sliding mode \cite{auvcontrolmethods}, and model predictive control \cite{auvmpc}. However, when considering how AUVs operate in environments that vary in space and time, control methods must be able to adapt to unpredictable changes in hydrodynamics. Furthermore, when deploying AUVs in the field, operators may wish to attach various payloads to the vehicle, shifting its physical parameters. On smaller AUVs, this can be especially problematic as payloads or small changes in environmental parameters (such as water density from salinity) can have large effects on the hydrodynamics (such as the center of buoyancy and mass) that are difficult to model and measure. 

Classical methods tend to require either significant and complex engineering to design robust and adaptive controllers, which are difficult to implement or transfer to new vehicles, or require time and labor-intensive manual tuning in each new configuration \cite{bhat2018hydrobatics}. Additionally, it is common to increase passive stability of the vehicle in these contexts by adding float to the top and weights on the bottom of the vehicle, but this trades of controllability and agility. Thus, a simpler method to design robust and adaptive controllers for AUVs is needed to support more streamlined or sophisticated field deployments. 

In this work we contribute: (1) a \textit{highly-parallelized, GPU-based AUV simulator} for designing and learning controls using reinforcement learning  methods \footnote{Code available at https://github.com/warplab/isaac-auv-env}, (2) to our knowledge, the first real-world demonstrations of a command-conditioned 6-DOF controller, direct to thrusters, deployed zero-shot on an AUV and (3) investigations of the sim-to-real deployment gaps on AUVs addressed by domain randomization and how it compares with hand-tuned PID controllers. We believe these strategies can reduce controller development effort and ease transferability to new systems and configurations.

%{In this work we contribute: (1) a highly-parallelized, GPU-based AUV simulator designed for learning robotic controls using reinforcement learning methods; (2) we also show how our simulator can be used to evaluate the robustness of RL-based controls to varying physical parameters; (3) finally, we demonstrate how controls learned in our simulator can be effectively deployed on a real vehicle with performance comparable to PID control. }

\begin{figure}[t]
  \centering
  \includegraphics[width=\columnwidth]{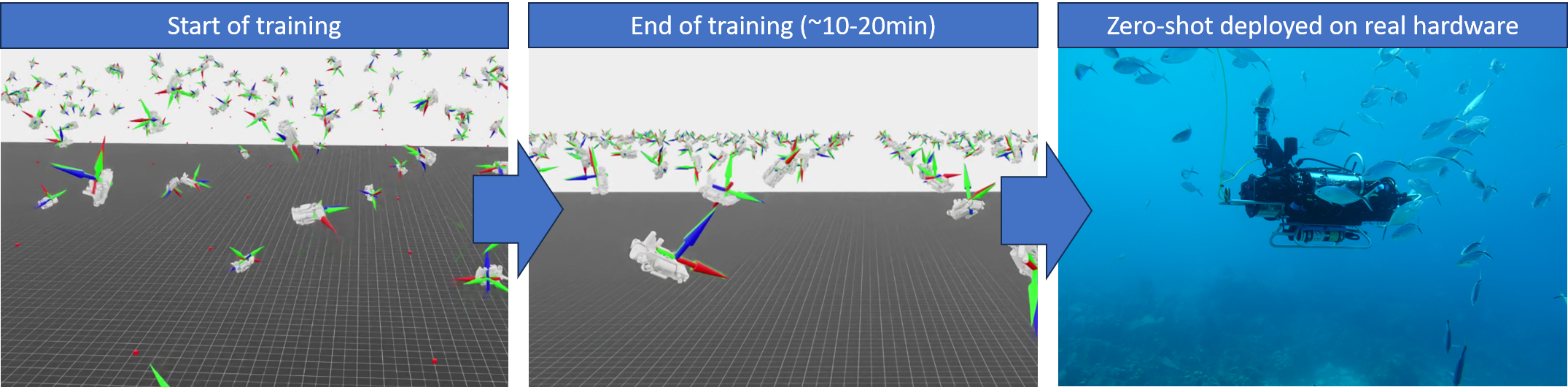}
  \caption{Using a highly parallelized training environment, we can rapidly train neural controllers that can take advantage of domain randomization techniques for robust zero-shot deployment on real hardware. For hardware evaluation, we use the CUREE vehicle \cite{curee}, shown here traveling through a coral reef environment interacting with the local wildlife.}
  \label{fig:overview}
\end{figure}

\section{RELATED WORK}

\subsection{Reinforcement Learning for Control}
Reinforcement learning (RL) methods for control of complex robotic systems has shown promising results ranging from applications in drones \cite{droneinwild,domainrandomizationdrones,droneracing}, legged robots \cite{leggedrobotrl,leggedrobotrl2,leggedrobotrl3}, and more \cite{rlcar}. It is especially appealing for its relatively simple implementations and robust outcomes in traditionally challenging control environments, and often can be applied across settings without in-situ tuning \cite{multireal}. 

\begin{figure}[h]
  \centering
  \includegraphics[width=\columnwidth]{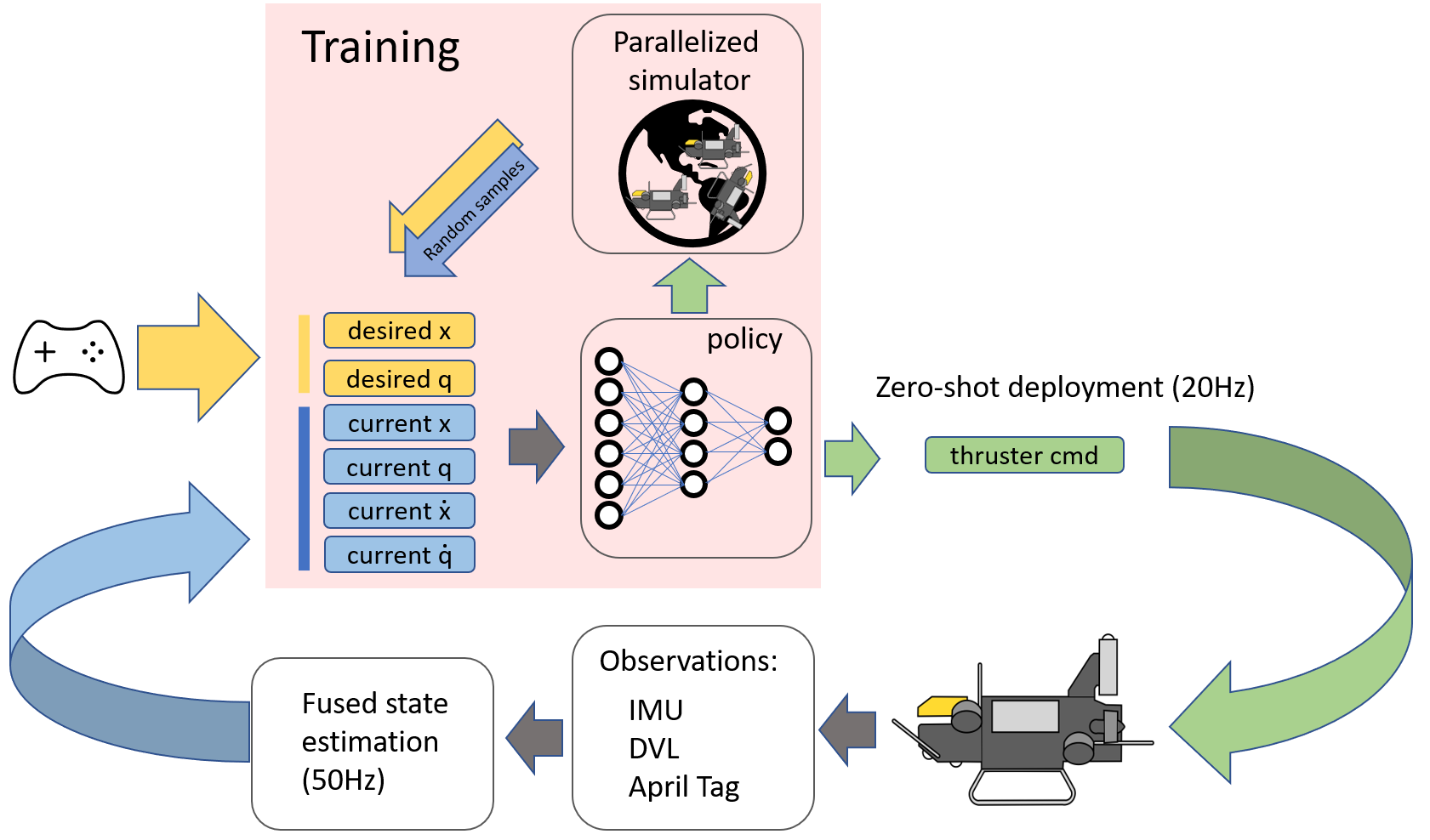}
  \caption{Our proposed approach uses on-policy reinforcement learning methods coupled with a highly parallelized underwater simulator to learn a robust control policy, which can be used directly on a robot with no additional training. This takes as input full 6-DOF commands relative to the AUV pose, and directly outputs thruster commands.}
  \label{fig:system}
\end{figure}

On-policy RL methods can often be trained efficiently in simulation, minimizing wear and tear on vehicles for training purposes \cite{learningtowalkparallel}. However, because simulation environments do not perfectly model the real-world, there is often a simulation-to-real performance (Sim2Real) gap. By leveraging a strategy known as \textit{domain randomization} (DR), which simulates uncertainty in dynamics and environmental parameters \cite{rldrtransfer} many systems are able to immediately apply to a variety of environmental contexts robustly and without tuning \cite{multireal}.

Multiple works have also demonstrated the effectiveness of using RL to control AUVs.  Meger et al. \cite{meger2015} were perhaps the first to demonstrate the use of a reinforcement learning-based approach to control a flipper based underwater vehicle. The approach used a Gaussian Process based model, learned from simulations and real world experiments, to predict the distribution of future states given the current state and control commands. The simulator used by the approach did not model hydrodynamic force, and was was slower than real-time, necessitating extensive real-world experiments to learn the control policy. 

Hadi et al. \cite{auvdrlpathplanning} explored the use of deep RL for learning 2-DOF control (yaw, speed) in a simulator. Masmitja et al. \cite{auvdrltracking} used RL to develop a tracking behavior for a surface vehicle, enabling it to follow underwater moving targets, but only focused on high-level steering commands. Lu et al.~\cite{ddr} is most similar to our approach, using RL with DR for sim-to-real transfer on an AUV, but does not have high-parallelization and limited to controls up to 4-DOF commands.

% This affords a significant amount of flexibility that reduces the amount of expert domain knowledge required to implement a controller for an AUV. RL problems are typically represented as Markov Decision Processes (MDP) where the robot's possible states and actions are represented by a stochastic process's state and action spaces respectively. The robot's controls are then represented by the process's policy $\pi$. RL methods can be classified as on-policy methods where the agent updates its behavior using data collected from its current policy or off-policy methods where data used in updating the policy is not necessarily the latest. The current state-of-the-art RL algorithm is called Proximal Policy Optimization (PPO) \cite{ppo}, a model-free, on-policy learning method. State-of-the-art off-policy methods include Soft Actor-Critic \cite{sac} and Dreamer \cite{dreamer}, both of which exceed PPO in sample efficiency.

\subsection{Simulators for Underwater Vehicles}
Realistic underwater simulations are necessary to mitigate the Sim2Real gap for AUVs. Gazebo \cite{gazebo, playergazebo} is a popular 3-dimensional robot simulator, but does not natively support underwater applications. UUV Simulator \cite{uuvsim}, and subsequently DAVE\cite{zhang2022dave}, introduce Gazebo plugins that enable large-scale, underwater simulations. However, Gazebo-based simulations are limited in their ability to produce high-fidelity renderings. Other works like UWSim \cite{uwsim} and UNavSim \cite{unavsim} make use of more graphics-oriented backends like OpenSceneGraph and Unreal Engine to produce more realistic renderings. These simulators also typically support ROS integration and enable simulation of a variety of real-world sensors such as Doppler Velocity Loggers (DVL) and LiDAR, enabling streamlined testing of AUV algorithms. MuJoCo \cite{mujoco} is a popular simulator that support parallelization across CPU cores, and has some native modeling of hydrodynamics though does not have underwater vehicle examples out of the box. However, of these approaches, Nvidia Isaac Labs, previously Isaac Gym, enables the simulation of thousands of robots in parallel through GPU-based optimizations. Furthermore, being built on top of NVIDIA's Isaac Sim framework, the simulator boasts high-quality visualizations. However, though it demonstrates success with terrestrial and aerial robots, an underwater environment has yet to be built with it \cite{isaacgym}. Our simulation approach is thus to combine the simplified hydrodynamics model presented in \cite{mujoco} with the GPU-based parallelization capabilities of \cite{isaaclab} to enable large-scale training environments for underwater vehicles.

\section{METHODS}
\begin{table}[thbp]
\caption{Default parameters for simulated AUV}
\label{tab:physicalparameters}
\begin{center}
\begin{tabular}{|c|c|c|}
\hline
$\rho$ & Water density & $997.0 \frac{kg}{m^3}$ \\
\hline
$\beta$ & Water viscosity & $0.001306 Pa\cdot s$ \\
\hline
$C_t$ & Thruster rotor constant & 0.001 \\
\hline
$B$ & COB-COM offset (m) & $[-0.05, 0.0, 0.01]$ \\
\hline
$V$ & Volume of AUV & $0.02275m^3$ \\
\hline
$\mathcal{M}$ & Mass of AUV & $22.701 kg$ \\
\hline
\end{tabular}
\end{center}
\end{table}

We are motivated by the problem of trying to learn a general 6-DOF controller for thruster-driven autonomous underwater vehicles, mapping full desired positional and orientation poses to low-level thruster commands, than can (a) be trained relatively quickly, (b) are relatively robust out of the box for sim-to-real transfer, with (c) minimal or no need for tuning for various payloads or hydrodynamic conditions. We note that (b) and (c) are similar aspects of \textit{sim-to-real domain transfer} problems.

Our methodology, naively shown in \Cref{fig:overview} and more detailed in \Cref{fig:system}, leverages highly parallelized simulations with on-policy reinforcement learning, which enables relatively fast learning from simulations, and \textit{domain randomization} (DR), similar to \cite{learningtowalkparallel,multireal}, which simultaneously allows for robust sim-to-real transfer and minimal tuning for various settings, but we apply these strategies to the underwater domain with a full 6-DOF capable controller.

In order to learn a policy that is robust to uncertainty in the AUV's physical parameters, we explore the application of domain randomization to the location of the center of buoyancy and AUV volume. The extent of the randomization we apply is shown in Table \ref{tab:dr_params}. Each environment is given 3 seconds to collect rewards during training. 

\begin{table*}[t]
\caption{Domain Randomization Configurations}
\label{tab:dr_params}
\begin{center}
\begin{tabular}{|c|c|c|}
\hline
Parameter & Distribution & \textcolor{red}{No DR}, \textcolor{green}{Small DR}, \textcolor{blue}{Large DR} \\
\hline
CoB-CoM Offset Noise & $UniformSphere(radius)$ & $radius = \textcolor{red}{0m}, \textcolor{green}{0.25m}, \textcolor{blue}{0.5m}$ \\
\hline
Volume Noise & $Uniform(range)$ & $range\approx \textcolor{red}{0L}, \textcolor{green}{1.5L}, \textcolor{blue}{3L}$ \\
\hline
\end{tabular}
\par\vspace{3mm}\par
\end{center}
\end{table*}

\subsection{GPU-based Simulation}

To better utilize domain randomization, in addition to exploring the large state space of a 6-DOF controller, we explore the use of GPU-based parallelization. We implement an underwater simulator with hydrodynamics in Nvidia Isaac Lab \cite{isaaclab}. This provides a pipeline for simulating up to thousands of agents simultaneously by leveraging the high parallelization capabilities of GPUs. Furthermore, it provides simple access to APIs and libraries for training models in a large variety of reinforcement learning settings that follow the OpenAI Gym \cite{openaigym} format.

\subsection{Simulated Hydrodynamic Model}

We model the hydrodynamic forces on the robot using the simplified inertia-based model presented in the physics engine MuJoCo \cite{mujoco} for drag, a separate model for buoyancy based on estimated volume, and a thruster model from \cite{uuvsim}. We describe these here for completeness. Intuitively, the simplified inertia-based model represents the vehicle as a rectangular prism and computes "equivalent" half side lengths $r_x, r_y, r_z$ from estimated inertial parameters $\mathcal{I}$. 

\begin{equation}
    r_i = \sqrt{\frac{3}{2\mathcal{M}}(\mathcal{I}_{jj}+\mathcal{I}_{kk}-\mathcal{I}_{ii})} \\
\end{equation}

where $\mathcal{M}$ is the mass of the body and $\mathcal{I}$ is its inertia matrix, which we naively estimate from the base-configuration of CUREE using a ruler and a scale (note that all subsequent deployments of CUREE have very different parameters to this due to payload changes which we do not have access).

Then, the fluid forces $\boldsymbol{f}_{inertia}$ and torques $\boldsymbol{g}_{inertia}$ are calculated as the sum of drag and viscous resistances which are calculated as

\begin{equation}
    \boldsymbol{f}_{\mathrm{inertia}}=\boldsymbol{f}_D + \boldsymbol{f}_V
\end{equation}

Each component of the drag terms is then calculated as

\begin{gather}
    f_{D,i}=-2\rho r_j r_k |v_i|v_i \\
    g_{D,i}=-\frac{1}{2}\rho r_i (r_j^4+r_k^4)|\omega_i|\omega_i
\end{gather}

And the viscous terms are calculated as 

\begin{gather}
    f_{V,i}=-6\beta \pi r_{eq} v_i \\
    g_{V,i}=-8\beta \pi r_{eq}^3 \omega_i
\end{gather}

where $r_{eq}=(r_x+r_y+r_z)/3$ is the equivalent radius for the drag on a spherical surface and $\beta$ is the fluid viscosity.

Buoyancy forces and torques are calculated using an estimated volume $V$ and center of buoyancy position $B$ which is specified relative to the center of mass. This separation is what we call the COB-COM offset.

For our thruster dynamics, we utilize a zero-order dynamic model for the thruster's angular velocity and convert its output to a thrust as proposed by Yoerger et al.~\cite{thrusterdynamics}.

\begin{equation}
    \mathrm{Thrust}(\Omega) = C_t \|\Omega\|\Omega,
\end{equation}
where $\Omega$ is the angular velocity of the thruster, and $C_t$ is a rotor constant. We derive PWM to angular velocity of the thruster based on datasheets from BlueRobotics on the T200, which is the thruster model used on CUREE.

The values of all physical parameters we use in our simulation are shown in Table \ref{tab:physicalparameters}.

\subsection{Learning-based controls for 6-DOF autonomous underwater vehicles}

\subsubsection{Learning environment and algorithm}

At a high-level, our approach is outlined in \Cref{fig:system}. In this work, we explore the use of a highly parallelized simulation environment coupled with an on-policy learning algorithm that is capable of quickly utilizing the experience gained. Furthermore, as proposed in \cite{multireal, learningtowalkparallel}, we utilize domain randomization, as in \cite{multireal}, in which various aspects of the simulated environment can be sampled randomly, to train a controller that is agnostic to those environmental changes (up to a point).

For the on-policy learning algorithm, we use the RSL-RL \cite{learningtowalkparallel} implementation of the Proximal Policy Optimization (PPO) algorithm \cite{ppo}.

\subsubsection{Observations, actions, and rewards}

We design a neural network-based policy, parameterized as a 2-layer MLP, that takes in an observation vector, with length of 17, $$\vec{o} = \{\vec{x}_{\text{offset}}, \vec{q}_{des}, \vec{q}, \vec{\dot{x}}, \vec{\omega} \}$$ where $\vec{x}_{\text{offset}}$ is the difference between the desired position and current position of the vehicle, in the vehicle's local frame of reference. As in \cite{learningtoflyseconds}, we learn a controller that drives this offset to the origin, which can then be applied for either position holding or trajectory following, while avoiding the need to sample the entire unbounded state space. We clip the offset error in deployments by the amount seen during training to avoid spurious behaviors and mitigates impacts of saturation. $\vec{q}$ are orientation quaternions and $\vec{\omega}$ are the angular velocities currently measured. The policy network outputs 6 thruster commands as actions, $\vec{a}$, that are normalized between \{-1, 1\}, which are scaled to PWMs. During training, environments are initialized at uniformly random across possible unit rotations and random offsets from respective origins of up to 2-meters, and are also provided random goal unit rotations. Resets occur at fixed step counts, equivalent to 3 seconds of experience time. We also use the guidance term presented in \cite{learningtoflyseconds} to speed up training.

We define the positive reward function at each time step $t$, which is the objective function the reinforcement learning algorithms attempts to maximize, as follows:

\begin{equation}
    r_t = \lambda_x r_x + \lambda_q r_q + \lambda_p r_p,
\end{equation}
where $\lambda_{\_}$ are the component weights; 
\begin{equation}
    r_x  = e^{ -\|x_\mathrm{offset}\|^2}
\end{equation}
is the reward associated with reaching the desired position;

\begin{equation} \label{eq:quatdistance}
    r_q = e^{- | \mathrm{angle}(\vec{q}_\mathrm{des} , \vec{q}) | }
\end{equation}
is the reward associated with robot orientation, and is measured in terms of the rotation angle of the difference quaternion $\vec{q}_\mathrm{des} \vec{q}^\star$, when expressed in axis-angle format. We use the provided method from \cite{isaaclab}. Here $q^\star$ represents the conjugate of the quaternion. 

\begin{equation}
    r_p = e^{- ||\vec{a}||^2 }
\end{equation}
is the reward associated with choosing control commands (actions) that minimize energy usage.

\begin{figure}[h]
  \centering
  \includegraphics[width=\columnwidth]{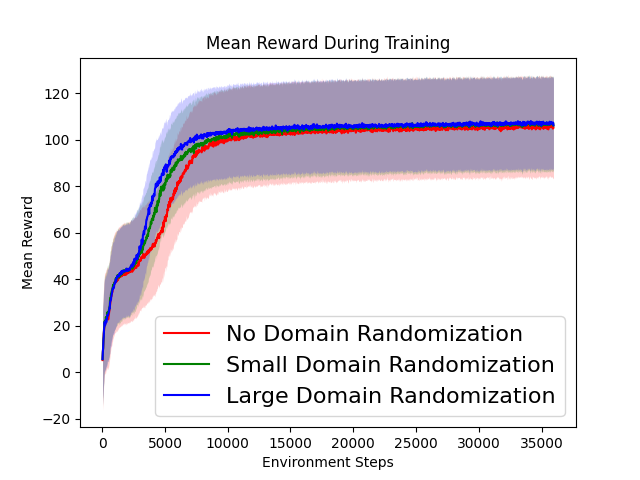}
  \caption{Mean reward collected during training with various amounts of domain randomization.}
  \label{fig:training}
\end{figure}

\begin{figure}[h]
     \centering
     \begin{subfigure}[b]{\columnwidth}
         \centering
         \includegraphics[width=\columnwidth]{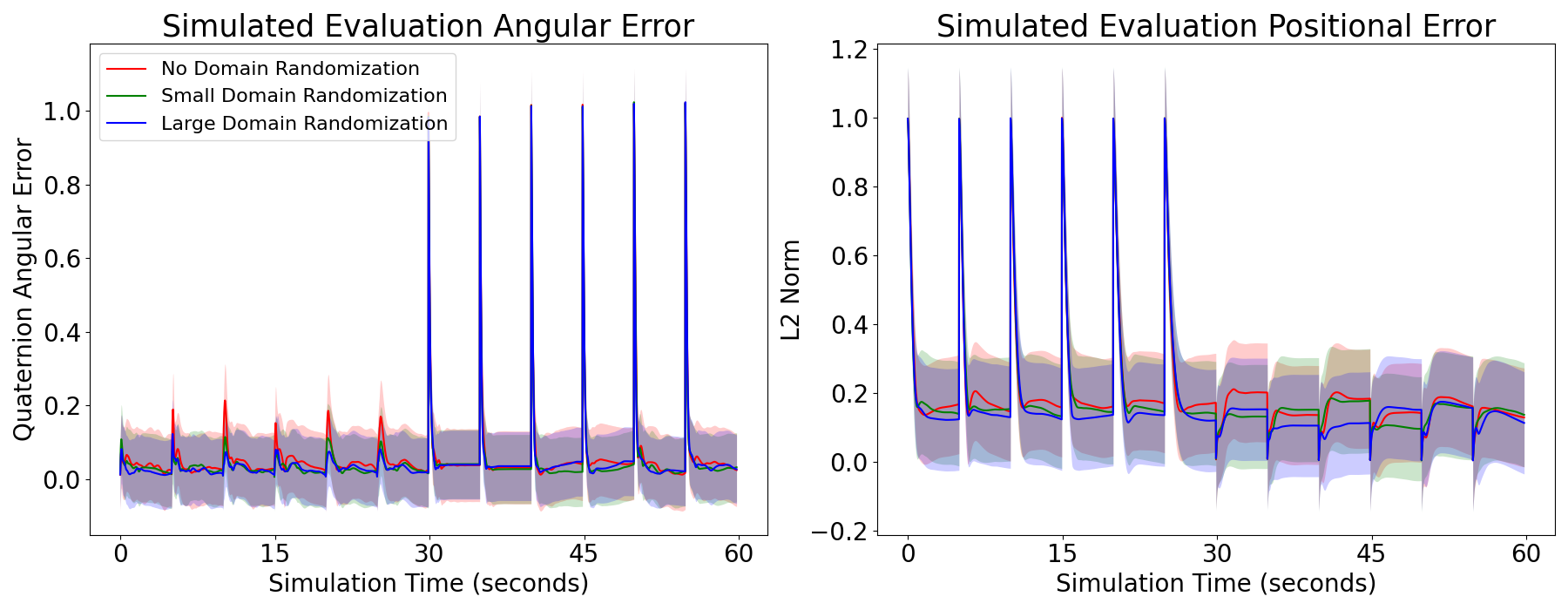}
         \caption{Ideal environment}
         \label{fig:sim2simeval1}
     \end{subfigure}
     \hfill
     \begin{subfigure}[b]{\columnwidth}
         \centering
         \includegraphics[width=\columnwidth]{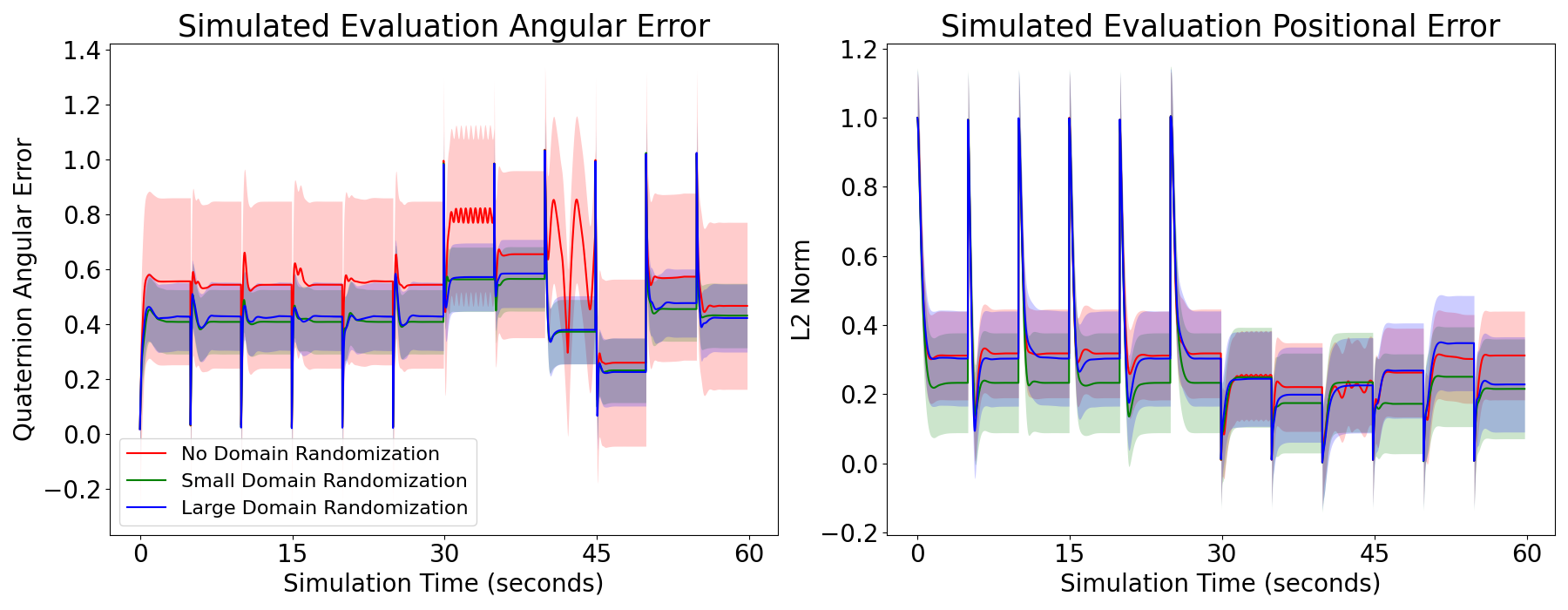}
         \caption{Lowered volume and small center of buoyancy shift}
         \label{fig:sim2simeval2}
     \end{subfigure}
    \caption{Plotted results from simulation testing of networks under different parameters demonstrating how networks trained using domain randomization exceed performance of naive networks when dynamic parameters are shifted. We use the $l_2$-norm and the quaternion distance described in Equation \ref{eq:quatdistance} to compute the errors.}
    \label{fig:sim2simevalplots}
\end{figure}

\begin{table}[h]
    \begin{subtable}[h]{0.45\textwidth}
        \centering
        \begin{tabular}{l | l | l | l}
         & No DR & Small DR & Large DR\\
        \hline \hline
        Angular MSE & 0.013 & \textbf{0.011} & \textbf{0.011}\\
        Positional MSE & 0.058 & 0.056 & \textbf{0.051}
       \end{tabular}
       \caption{Ideal environment}
       \label{tab:sim2simeval1}
    \end{subtable}
    \hfill
    \begin{subtable}[h]{0.45\textwidth}
        \centering
        \begin{tabular}{l | l | l | l}
         & No DR & Small DR & Large DR\\
        \hline \hline
        Angular MSE & 0.401 & \textbf{0.195} & 0.204\\
        Positional MSE & 0.108 & \textbf{0.081} & 0.104
       \end{tabular}
       \caption{Lowered volume and small center of buoyancy shift}
       \label{tab:sim2simeval2}
    \end{subtable}
     \caption{Evaluation results from simulation comparing performance of networks trained with and without domain randomization under various dynamics parameters.}
     \label{tab:sim2simeval}
\end{table}

\subsubsection{Domain transfer (validation in simulation)}

Domain randomization has been shown to be an effective strategy for domain transfer, both in purely simulation and sim-to-real applications, as discussed in \cite{multireal}. We focus our analysis on issues that are more unique and reflective of the issues that one may encounter in the underwater domain, compared to that of quadrotors \cite{multireal} or quadrupeds \cite{learningtowalkparallel}. In particular, it is common in AUVs to add new underwater housings for various deployments, usually equipping the vehicle with different sensing or computing capabilities. These additional housings will typically add both \textit{volume} and hence potentially shift the \textit{COB-COM} offset, which acts as a stabilizing force (between gravity and buoyancy) for underwater vehicles.

To do this, we train 3 policy networks with varying degrees of domain randomization, from none, medium, and large, on the volume and COB-COM offset terms, as specified in \Cref{tab:dr_params}, these environments are sampled around the default settings in \Cref{tab:physicalparameters}.

To test the robustness to domain transfer, we evaluate each policy against an environment with the default settings and another with shifted volume and COB-COM offset that is commanded to move in both directions along each of the 3 linear and 3 orientation axes by a fixed distance or angle (such as 1-meter forward or 60-degrees yaw), and given roughly 5 seconds to do so. The best performing policy across both domains is selected for testing in the real-world.

\begin{figure*}[h]
    \centering
    \includegraphics[width=\textwidth]{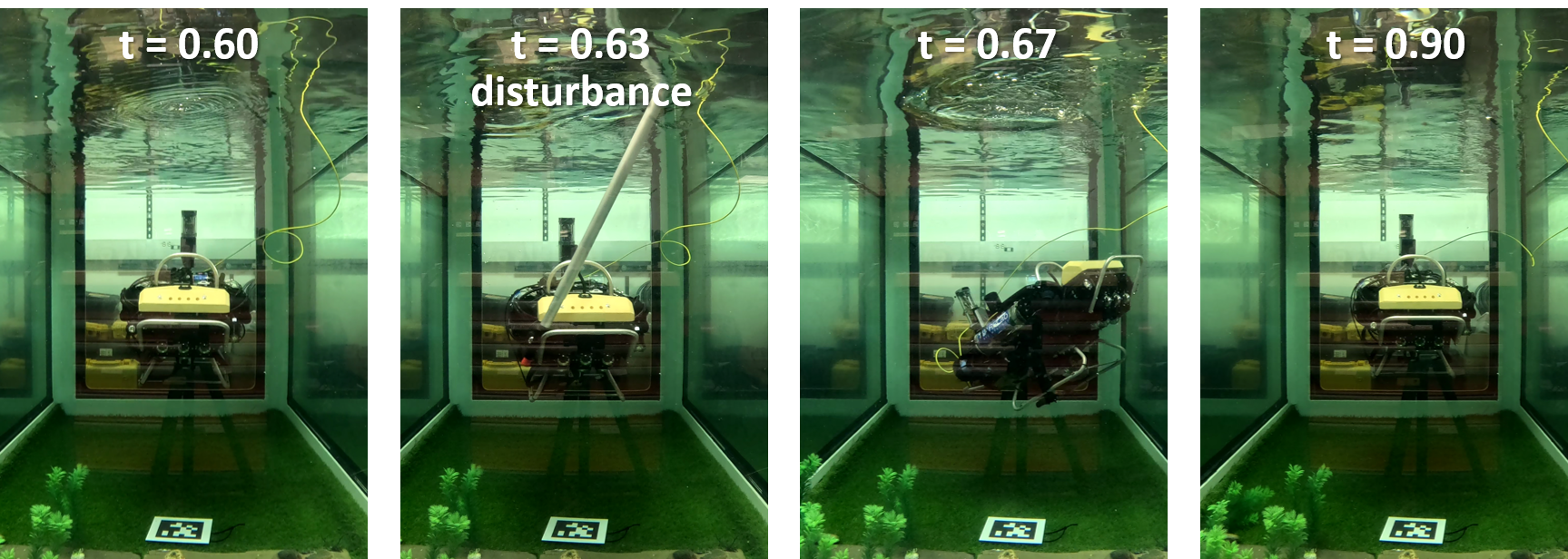}
    \caption{Real world testing experiments, showing neural controller ability to reject large disturbances while holding position and orientation. Note that the vehicle is ballasted to be positively buoyant, so must actively position hold. The controller is running on-board the vehicle's Jetson Orin NX and sending low-level motor commands at 20Hz. A human with a stick drags CUREE aggressively to the side, and the neural controller is able to return. An AprilTag is used for vision-based feedback to provide global position in the tank. This is fused with DVL and IMU feedback for faster position feedback and for when the AprilTag can no longer be seen. Timestamps are in minutes and manually aligned with \Cref{fig:neural-disturb}.}
    \label{fig:neural-disturb-qual}
\end{figure*}

\begin{figure}[h]
    \centering
    \includegraphics[width=\columnwidth]{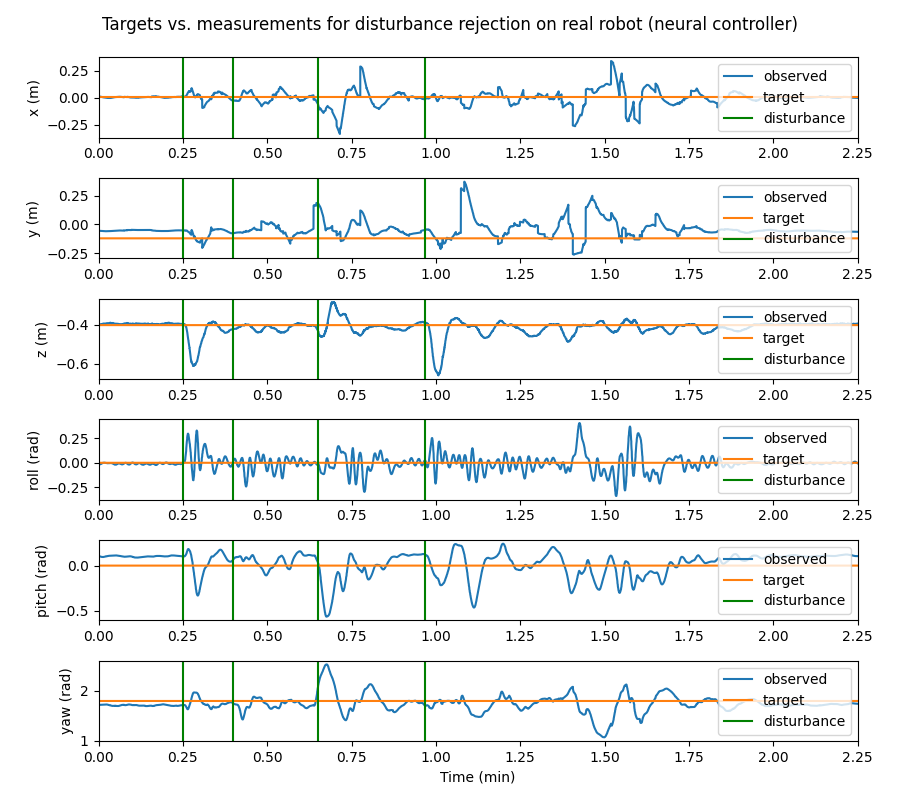}
    \caption{Results of disturbance rejection experiment when utilizing a neural controller trained using the described RL strategy. The AUV is set to position hold at the target setpoints while a large external impulse force (stick pushes the robot briefly) is applied. Disturbances are marked by green vertical lines (manually annotated through video review). First disturbance occurs at 0.25 min.}
    \label{fig:neural-disturb}
\end{figure}

\begin{figure}[h]
    \centering
    \includegraphics[width=\columnwidth]{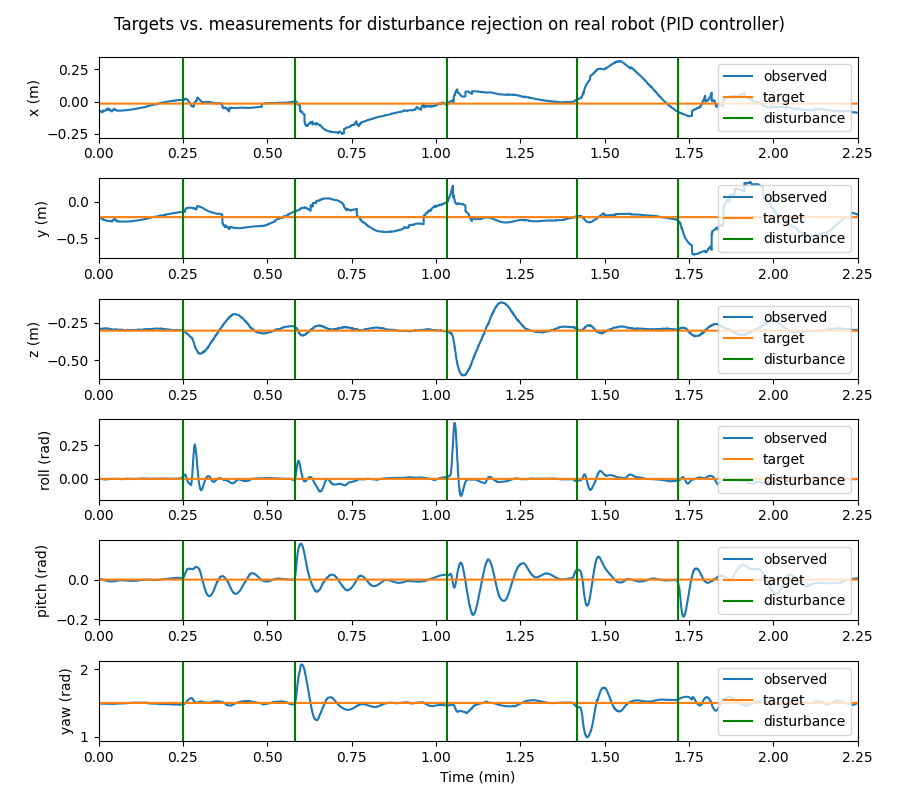}
    \caption{Results of disturbance rejection experiment when utilizing a PID controller, naively and manually tuned. Disturbance methodology is the same as in \Cref{fig:neural-disturb} and the first disturbance occurs at 0.25 min.}
    \label{fig:pid-disturb}
\end{figure}

\subsection{Domain transfer (sim-to-real)}

To test the transfer-ability of the trained network, we deploy our selected policy from simulation onto CUREE \cite{curee}. CUREE is an AUV equipped with 6 BlueRobotics T200 thrusters for 6-DOF movement and a Nvidia Jetson Orin NX, in a water-cooled housing, that allows us to run our neural network policy on-board in real-time ($\sim$20Hz). On-board cameras (which are used to detect AprilTags), a Doppler Velocity Logger (DVL) and an Inertial Measurement Unit (IMU) are fused via an EKF to provide position, orientation, and velocity estimates in the body frame \Cref{fig:system}. The sensors and thrusters are all controlled and monitored through a Raspberry Pi 4, and everything is implemented in ROS for coordination. These attributes make CUREE an ideal platform for testing our RL-based controls. 

The policy takes state estimates from the EKF and a desired pose from a user or trajectory, and outputs thruster commands. We note that unlike in \cite{learningtoflyseconds,margolis2024rapid}, we do not model motor action delays which can lead to instabilities in deployment. To account for un-modeled delays, we apply a quaternion slerp to smooth out large jumps between the desired and estimated quaternions for stable sim-to-real transfer. To test the robustness and accuracy of the controller, CUREE is commanded to hold a position above an AprilTag while it is aggressively perturbed via external forces. This setup is shown in \Cref{fig:neural-disturb-qual}.

As we do not introduce our agent to any data collected in the real-world during the training process, our method is fully zero-shot.

\section{Results and Discussion}
\subsection{Training}

The mean reward curves across the different training configurations are shown in \Cref{fig:training}. We are able to simulate 2048 environments in parallel while using roughly 11.4 GB of GPU memory on-board an Nvidia A6000 GPU. Total training time was roughly 10-20 minutes. We note that after the initial spike from running the base simulation, increasing the number of environments led to very minimal increase GPU memory usage, so we theorize that we could run simulate significantly more environments at a time. Counter-intuitively, we observe that using domain randomization increases the convergence rate of the training.

\subsection{Evaluating domain transfer in simulation}

The results of our simulation-based domain transfer evaluation are in \Cref{fig:sim2simevalplots}, which are accumulated into the results in \Cref{tab:sim2simeval}. The ideal setting is simply testing the trained controller on the nominal environment in \Cref{tab:physicalparameters}, while the simulated domain shifted vehicle has a COB-COM that is 0.2 meters in front of the vehicle's COM, and with a 1.5L lower volume (so it is negatively buoyant). 

We see that some domain randomization provides improved domain transfer performance, but there is a limit, as further domain randomization causes decreasing performance at a point. This has also been discussed in \cite{multireal}, but intuitively, can be hypothesized that the domain randomization process is only learning an averaged behavior across domains, rather than one that can adapt to maximize performance in each. At some point, this results in too restrictive/conservative of a controller. We thus select the small-DR trained policy for testing in the real world.

\subsection{Sim-to-real transfer}

We test zero-shot transfer of the policy directly on real hardware in a small tank environment. We are deploying on a vehicle with relatively unknown hydrodynamics, as it contains an arbitrary sensor payload configuration and we do not measure the new center of mass or buoyancy. By recording both the commanded poses and estimated poses we are able to evaluate how closely CUREE is able to hold a position and how it recovers from aggressive disturbances as shown in \Cref{fig:neural-disturb} and \Cref{fig:neural-disturb-qual}. These show that CUREE, with the trained neural network policy, running at roughly 20Hz, is able to reject aggressive disturbances. This is rather surprising given the simplistic simulated modeling and domain randomization. We note around the 1.4-min mark, an error causes an unplanned disturbance, but the vehicle returns to steady state. When compared to the performance of the PID controller \Cref{fig:pid-disturb}, the neural controller seems more aggressive, which aligns with findings in similar studies in quadrotors \cite{multireal,learningtoflyseconds}.

Additionally, it is interesting to note the steady-state error that is present in the neural controller, especially in the pitch and y-directions in \Cref{fig:neural-disturb}. This is because the controller only learns a conservative policy that works well on average, but cannot adapt to specific deployment conditions. In the future, online learning and adaptation techniques such as those presented in \cite{margolis2024rapid} can help alleviate these types of errors.

Further errors that we noticed in deployments that are not captured by our model are issues such as major thruster imbalances (housings that do not have streamlined geometry severely hinder thruster performance in asymmetric ways).

\section{Conclusion}
% \vspace{-10pt}
In this work, to our knowledge, we provide the first implementation of a highly parallelized simulation for underwater vehicle control that can be used in reinforcement learning pipelines. We use it to train, in minutes, a control policy that maps high-level 6-DOF commands directly into thruster allocations. We use domain randomization methods to address sim-to-real robustness across various dynamic settings, allowing deployments with no additional tuning at test-time. We validate this by deploying a trained policy zero-shot, running on-board in real-time a custom AUV that can recover from aggressive disturbances, with no a priori information about precise physical AUV parameters such as centers of mass and buoyancy. This approach opens the door for rapid experimentation with various other AUV configurations, without the need to design individual controllers for each.
    
In the future we plan to explore using online learning and adaptation techniques such as those proposed by \cite{animalstyle,margolis2024rapid,dreamer,daydreamer} to develop a controller that can adapt to new physical circumstances in real-time, and resolve issues such as the steady state errors we observed. Finally, we hope to integrate other types of hydrodynamics models to enable actuation methods outside of thrusters, which are substantially more difficult to develop controllers due to their highly nonlinear nature.

% \addtolength{\textheight}{-12cm}   % This command serves to balance the column lengths
                                  % on the last page of the document manually. It shortens
                                  % the textheight of the last page by a suitable amount.
                                  % This command does not take effect until the next page
                                  % so it should come on the page before the last. Make
                                  % sure that you do not shorten the textheight too much.

%%%%%%%%%%%%%%%%%%%%%%%%%%%%%%%%%%%%%%%%%%%%%%%%%%%%%%%%%%%%%%%%%%%%%%%%%%%%%%%%

%%%%%%%%%%%%%%%%%%%%%%%%%%%%%%%%%%%%%%%%%%%%%%%%%%%%%%%%%%%%%%%%%%%%%%%%%%%%%%%%

%%%%%%%%%%%%%%%%%%%%%%%%%%%%%%%%%%%%%%%%%%%%%%%%%%%%%%%%%%%%%%%%%%%%%%%%%%%%%%%%

\section*{ACKNOWLEDGMENTS}
This work was supported in part by WHOI Summer Student Fellowship Program, WHOI Independent Study Award, WHOI Academic Programs Office Endowed Funds, and NSF NRI Award No. 2133029.
Thanks to Ethan Fahnestock for the initial simulator and learning pipeline. Thanks also to Nathan McGuire, Peter Werner, Daniel Yang, and Tim Seyde for their help and discussions.

% \begin{thebibliography}{99}

% \bibitem{c3} H. Poor, An Introduction to Signal Detection and Estimation.   New York: Springer-Verlag, 1985, ch. 4.
% \bibitem{c4} H. Poor, An Introduction to Signal Detection and Estimation.   New York: Springer-Verlag, 1985, ch. 4.

% \end{thebibliography}

% \bibliographystyle{plain}
% \bibliography{refs}

{\small
\bibliographystyle{IEEEtran}
\bibliography{refs}
}

\end{document}